%% file: main.tex
\documentclass{article}
\usepackage[preprint]{neurips_2026}
\input{config}

\definecolor{EasyBestBg}{RGB}{222,235,255}
\definecolor{MediumBestBg}{RGB}{224,242,228}
\definecolor{HardBestBg}{RGB}{255,235,214}

\providecommand{\easybest}[1]{#1}
\providecommand{\mediumbest}[1]{#1}
\providecommand{\hardbest}[1]{#1}

\renewcommand{\easybest}[1]{\cellcolor{EasyBestBg}\textbf{#1}}
\renewcommand{\mediumbest}[1]{\cellcolor{MediumBestBg}\textbf{#1}}
\renewcommand{\hardbest}[1]{\cellcolor{HardBestBg}\textbf{#1}}

\title{SONG: A Photorealistic 3D Gaussian Simulation Platform for Benchmarking Social Navigation}

\author{%
    Weiqi Huang$^{1,*}$, Dianyi Yang$^{1,*}$, Jiaxin Li$^{1}$, Shuangyi Dong$^{1}$, Hao Xu$^{1}$, Zan Wang$^{1}$, Wei Liang$^{1\,\textrm{\Letter}}$
    \vspace{6pt}\\
    \small $^1$ School of Computer Science \& Technology, Beijing Institute of Technology
    \vspace{6pt}\\
}

\begin{document}

\maketitle
\begin{figure}[ht!]
    \centering
    \includegraphics[width=\linewidth]{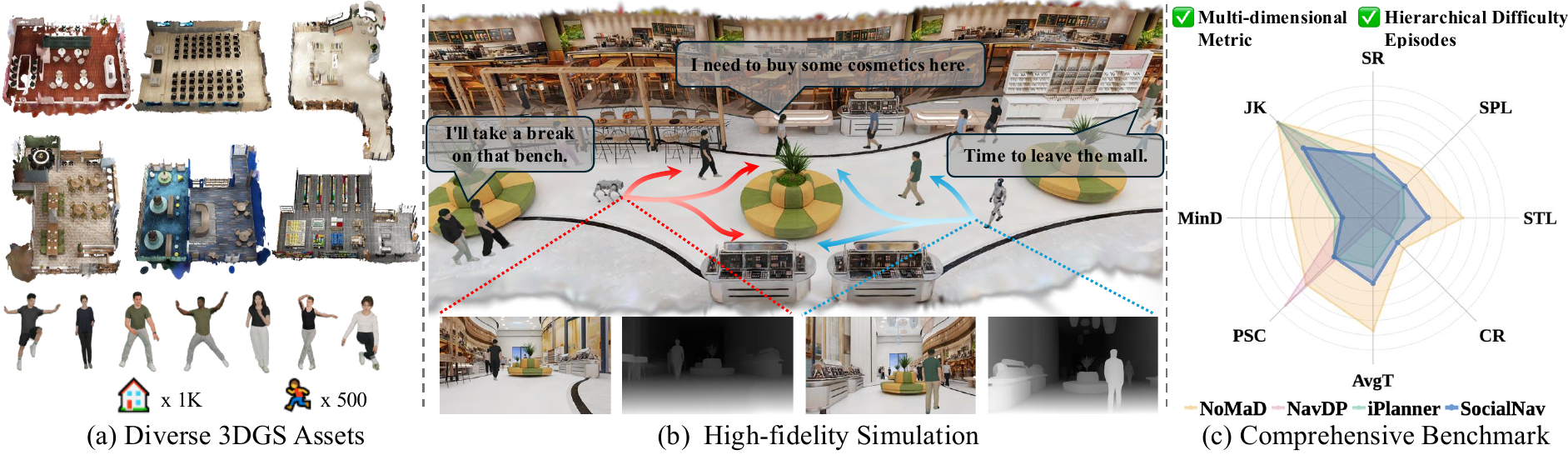}
    \caption{\textbf{\platform Overview.} We introduce a high-fidelity benchmark for social navigation. (a) Our asset library comprises 1,000 photorealistic 3DGS environments and 500 3DGS dynamic human avatars. (b) The simulation engine generates reasonable social episodes with tiered complexity, providing multi-modal first-person perceptions.  (c) A holistic evaluation framework benchmarks SOTA policies across effectiveness, safety, and social compliance metrics.}
    \label{fig:teaser}
\end{figure}

\begin{abstract}
Social navigation has progressed from simplified 2D environments toward a more general vision-based setting, in which a robot needs to achieve socially compliant behavior purely from onboard visual observations. 
Yet supporting simulation platforms have not kept pace: existing options either lack visual observations, lack moving human avatars, or fall short of real-world fidelity in appearance and pedestrian behavior, offering limited support for advancing vision-based social navigation.
We introduce \platform, a \textbf{SO}cial \textbf{N}avigation platform powered by 3D \textbf{G}aussian splatting (3DGS). It leverages 3DGS for both scene and avatar representations, drives pedestrians using semantically grounded trajectories generated by a large language model, and synthesizes their full-body motion with a trajectory-conditioned generator to produce continuous, natural movement. On top of the platform, we curate \benchname, a set of evaluation episodes stratified by difficulty, and propose a multi-dimensional metric suite covering effectiveness, safety, and social compliance. A systematic evaluation of representative navigation baselines reveals three findings: (a) vision-based social navigation is far from solved; (b) a critical safety deficit precedes social etiquette; (c) real-world data matters more than model scale. Crucially, we demonstrate that fine-tuning on our curated data effectively improves the success rate in real-world environments. We hope our platform provides a faithful and rigorous testbed for the next generation of vision-based social navigation research.
\end{abstract}

\section{Introduction}
Autonomous robot navigation has undergone extensive investigation across a diverse spectrum of tasks and objectives, ranging from various goal-directed paradigms \cite{anderson2018evaluation, wijmans2020ddppo, chaplot2020object, krantz2023navigating, wang2022towards} and vision-language navigation (VLN) \cite{anderson2018vision, krantz2020beyond, wang2025rethinking} to autonomous exploration \cite{zhu2025move} and floorplan-guided navigation \cite{chen2024f3loc, li2024flona, huang2025floor}. While these advancements have significantly improved a robot's ability to reason about spatial layouts and goal-directed semantics, they often overlook the necessity for robots to operate effectively within real-world environments dense with dynamic human activities. In such human-populated spaces, robots must navigate in a socially compliant manner: respecting personal space, avoiding disruptive maneuvers, and remaining predictable to nearby humans. These demands, which go beyond the classical objectives of efficiency and safety, define the problem of social navigation. As the problem has drawn increasing attention \cite{francis2025principles, han2025dr, liu2025height,liu2021decentralized,liu2022intention, han2025ratatouille,gong2025cognition, chen2511socialnav, hu2026navthinker}, the field has gradually moved away from early formulations that relied on simplified 2D environments and strong assumptions about pedestrian states, toward a more general vision-based setting, in which the robot needs to achieve socially compliant behavior purely from onboard visual observations.

Despite this task evolution, existing simulation platforms have lagged behind, falling broadly into three categories that inadequately support vision-based social navigation. (i) 2D simulators \cite{chen2019crowd, biswas2022SocNav, Han_IR-SIM_An_Open-Source_2026} reduce the environment to top-down geometry, where the policy never sees a pixel. (ii) 3D platforms \cite{shen2021igibson, savva2019habitat} supporting static navigation offer rich visual observations but lack dynamic avatar agents, often collapsing the task into semantic goal finding. (iii) As detailed in \cref{tab:simulator_comparison}, the handful of 3D platforms that do model pedestrians still fall short of real-world fidelity. Specifically, mesh-based rendering leaves a visible appearance gap; pedestrians follow hand-crafted rules toward random waypoints without semantic intent; and motions assembled from looping mocap clips produce stiff and repetitive gaits.

To address this gap, we introduce \platform, a high-fidelity platform for benchmarking vision based social navigation, built around three coupled designs, as shown in \cref{fig:teaser}.
\begin{itemize}[leftmargin=*]
    \item \textbf{Photorealistic 3D Gaussian assets}. To overcome the visual fidelity limitations of mesh based benchmarks, we curate a library of 500 diverse human Gaussian avatars. This collection combines 195 subjects from DNA Rendering \cite{cheng2023dnarendering} for fine grained structural details with 305 image conditioned reconstructions via LHM \cite{qiu2025lhm} to ensure broad appearance diversity. These agents are seamlessly integrated into photorealistic static scenes from SAGE-3D \cite{miao2025sage3d}, all managed within a unified, real time Gaussian rendering framework. This cohesive environment effectively bridges the visual sim-to-real gap, providing vision-centric agents with unprecedented sensory fidelity.
    \item \textbf{Semantics-Driven Social Dynamics}. Moving beyond randomized pedestrian movement, we propose a hierarchical behavior engine that synthesizes intent-aware pedestrian dynamics. We leverage a large language model to reason over 2D semantic maps, generating long-horizon activity sequences (e.g., navigating from reception to a workstation) that serve as high-level intents. These intents are grounded in multi-agent trajectories through a joint crowd planner and subsequently embodied via Kimodo \cite{rempe2026kimodo}. By binding fluid full-body motions to our Gaussian avatar library, this pipeline replaces stiff mocap loops with natural, semantically-consistent gaits.
    \item \textbf{Rigorous Evaluation Protocol}. We develop an automated pipeline to curate \benchname, a benchmark consisting of 500 diverse evaluation episodes. These episodes are systematically stratified into three difficulty levels based on path length and crowd density. To ensure a multi-dimensional assessment, we define a metric suite across three pillars: effectiveness, safety, and social compliance. We also provide human teleoperated trajectories as empirical upper bounds to accurately measure the performance gap of current vision based baselines.
\end{itemize}

Building on \benchname, we conduct a systematic evaluation of representative navigation methods, surfacing three findings:
\textbf{(a) Vision based social navigation is far from solved.} All baselines exhibit a severe performance drop on \benchname, with success rates remaining below 22\% in easy settings and plummeting to near zero in hard ones, revealing that current models still fail to navigate effectively in realistic social scenarios.
\textbf{(b) Safety deficits precede social etiquette.} Before considering complex social rules, current models require significant improvements in basic safety. High failure rates persist even in avatar-free scenarios, revealing a fundamental difficulty in mapping complex observations to safe geometric spaces. Furthermore, agents lack collision recovery mechanisms, typically remaining stuck after contact rather than replanning.
\textbf{(c) data realism trumps model scale}. Beyond these diagnostics, we evaluate the real-world generalization of \platform. Finetuning on our curated multi modal data boosts the success rate of a Unitree Go2 robot, validating the physical world fidelity of our simulation assets.

Our contributions are threefold:
\begin{itemize}[leftmargin=*]
    \item \textbf{Photorealistic Social Navigation Platform.} We develop a high fidelity simulation environment that integrates 3DGS based photorealistic rendering with LLM generated pedestrian intent and Kimodo driven motion. This platform includes a complete toolchain for scene authoring, pedestrian scripting, and autonomous robot data collection.
    \item \textbf{Comprehensive Benchmark.} We establish a rigorous benchmarking protocol featuring curated episodes across multiple scenes, stratified by interaction complexity. Our multi-dimensional evaluation suite covers efficiency, safety, and social compliance.
    \item \textbf{Systematic Evaluation and Analysis.} We conduct a comprehensive study of representative navigation methods on \benchname. Our analysis reveals critical performance gaps in high fidelity environments and provides fundamental insights into the generalization bottlenecks of current learning paradigms. Furthermore, real-world experiments on a Unitree Go2 robot validate the effectiveness of our data.
\end{itemize} 

\begin{table}[t]
    \centering
    \caption{\textbf{Comparison of social navigation platforms.} \platform uniquely combines photorealistic rendering, semantically plausible avatar trajectories, and fluid, natural avatar motion. Repr.: Representation; Anim.: Animation.}
    \label{tab:simulator_comparison}
    \resizebox{\linewidth}{!}{%
        \begin{tabular}{lcccccc}%
            \toprule
            \multirow{2}{*}{\textbf{Platform}} 
                & \textbf{Num.}      & \textbf{Scene}          & \textbf{Num.}     & \textbf{Avatar}   & \textbf{Avatar}              & \textbf{Motion} \\
                & \textbf{Scenes}    & \textbf{(Source/Repr.)} & \textbf{Avatars}  & \textbf{Repr.}   & \textbf{Trajectory}         & \textbf{Source} \\
            \midrule
            iGibson \cite{shen2021igibson}          & 15   & Real/Mesh   & 3   & Mesh & Random             & No Anim. \\
            HabiCrowd \cite{vuong2024habicrowd}     & 480  & Real/Mesh   & 40  & Mesh & Manual Goal        & No Anim.  \\
            Habitat 3.0 \cite{puig2023habitat3}     & 59   & Syn./Mesh   & 12  & Mesh & Manual Goal        & MoCap Anim.            \\
            Social-HM3D \cite{gong2025cognition}    & 844  & Real/Mesh   & 12  & Mesh & Manual Goal        & MoCap Anim.            \\
            \midrule
            \textbf{SONG (Ours)}                   & \textbf{1000} & \textbf{Real/3DGS} & \textbf{500} & \textbf{3DGS} & \textbf{LLM Semantic Goal} & \textbf{Generative (Traj.+Text)} \\
            \bottomrule
        \end{tabular}%
    }%
\end{table}

\section{Related work}
\subsection{Social navigation}
Social navigation, which entails safe and efficient robot movement through human-populated spaces, has remained a cornerstone of robotics research for decades \cite{fox2002dynamic,mavrogiannis2023core}. Earlier paradigms like SFM \cite{helbing1995social} and ORCA \cite{van2011reciprocal} rely on rigid rules and perfect state perception, often leading to the freezing robot problem. To handle more general settings, research has shifted toward learning-based approaches, categorized into two primary paradigms: modular pipelines and end-to-end vision-based methods. Modular methods \cite{han2025dr, han2025ratatouille, liu2021decentralized, liu2022intention, liu2025height} decouple the task by first extracting pedestrian positions via a detector. These spatial coordinates, together with the robot's proprioceptive state and target location, are then fed into a policy to compute navigation actions. However, these frameworks remain vulnerable to noisy perception and detection errors that propagate through the pipeline. In contrast, end-to-end vision-based methods \cite{gong2025cognition, hu2026navthinker, chen2511socialnav} learn a direct mapping from raw visual inputs to control outputs, bypassing explicit human localization. While this paradigm mirrors human-like navigation, its progress is bottlenecked by the prohibitive cost of real-world data and the substantial sim-to-real gap in existing simulators, which often lack the photorealism and sophisticated human behaviors necessary for vision-centric policies. To bridge this gap, we introduce \platform, a photorealistic platform designed to provide a faithful testbed for vision-based social navigation.

\subsection{Navigation platform}
Early social navigation platforms operate in 2D environments \cite{chen2019crowd, biswas2022SocNav, Han_IR-SIM_An_Open-Source_2026}, where pedestrians are abstracted as discs driven by rule-based planners such as ORCA \cite{van2011reciprocal} or the SFM \cite{helbing1995social}, or replayed from real-world trajectory datasets \cite{biswas2022SocNav}. Although these lightweight environments are well-suited for benchmarking modular navigation pipelines, the absence of any visual observation makes them fundamentally incompatible with vision-based policy learning. iGibson \cite{shen2021igibson, igibsonchallenge2021} enables RGB and depth rendering in reconstructed 3D scenes, with pedestrians driven by ORCA but represented as rigid, unanimated avatars. Habitat 3.0 \cite{puig2023habitat3} improves human avatar fidelity by introducing articulated SMPL-X humanoids with diverse body shapes and motions. Despite these improvements, mesh-based rendering across both platforms produces synthetic appearances that diverge substantially from real camera observations, imposing a photometric domain gap that undermines sim2real transfer. The Arena series \cite{kastnerarena, kastner2024arena, shcherbyna2025arena}, while spanning 2D to 3D backends with configurable social force models and rich benchmarking tooling, similarly relies on low-fidelity mesh avatars with rule-based motion, leaving the visual realism gap unaddressed.

Our \platform closes these gaps by reconstructing both scenes and human avatars via Gaussian Splatting, achieving photorealistic rendering that faithfully mirrors real-world appearances. Pedestrian trajectories are generated by an LLM-driven behavior model for socially plausible interactions, and body animations are synthesized via Kimodo for smooth, natural motion—collectively providing a faithful testbed for vision-based social navigation.

\section{\platform Platform}
\label{sec:socialgs_platform}

\subsection{Overview}
\label{sec:socialgs_overview}

\begin{figure}[t]
    \centering
    \includegraphics[width=\linewidth]{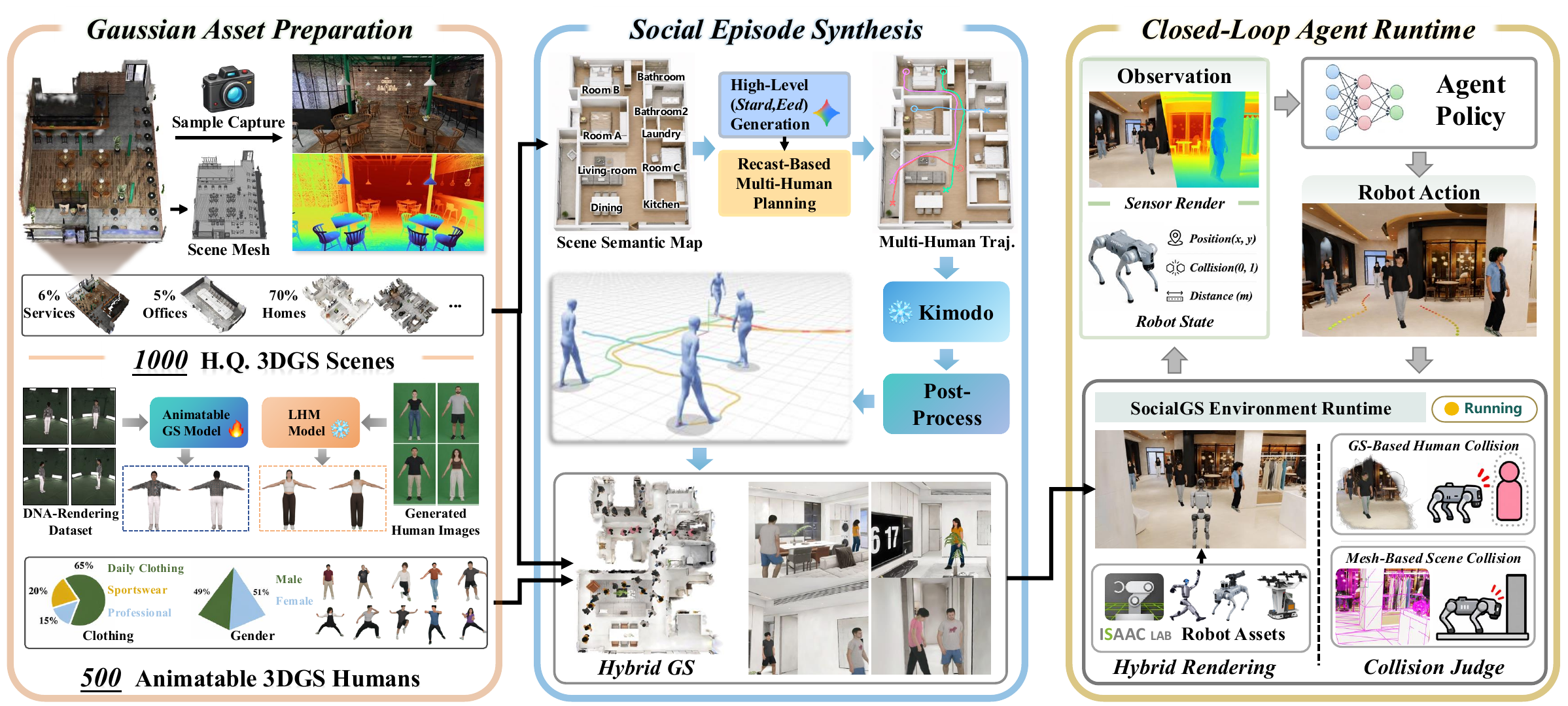}
    \caption{\textbf{\platform Pipeline.}  The platform prepares Gaussian scene and avatar assets, synthesizes social episodes from semantic intents, trajectories, motions, and avatar bindings, and exposes the resulting dynamic Gaussian scene through a closed-loop runtime for rendering and collision-aware agent interaction.}
    \label{fig:socialgs_pipeline}
\end{figure}

In this work, we introduce \platform, a high-fidelity platform specifically engineered to advance vision-based social navigation. The platform is distinguished by three key attributes: 
(1) \textbf{High-Fidelity Rendering}: both static environments and dynamic humans are represented with 3D Gaussian Splatting, providing visually consistent and high-quality observations;
(2) \textbf{Socially Realistic Behaviors}: pedestrian trajectories are synthesized and validated by Large Language Models (LLMs), while avatar motions are generated via Kimodo and refined manually to ensure both fluid motion and logical social interaction; 
(3) \textbf{Extensibility}: the platform provides versatile interfaces that empower users to easily customize scenes, pedestrian profiles, and navigation tasks. As illustrated in \cref{fig:socialgs_pipeline}, \platform functions through three core stages, each of which is detailed below.

\subsection{Gaussian asset preparation}
\label{sec:socialgs_assets}

The first stage prepares the static scene assets and animatable avatar assets used by \platform. For scenes, we adopt SAGE-3D-style assets, where each 3D Gaussian Splatting scene~\cite{kerbl2023gaussian} is paired with semantic and navigation-oriented metadata~\cite{miao2025sage3d}. Each scene contains a Gaussian representation for photorealistic rendering, a lightweight geometry proxy for simulator interoperability, and a top-down semantic/traversability map for activity sampling and feasibility checking. The semantic map is used only during episode construction, while agent observations are rendered directly from the Gaussian scene. All representations are registered in a shared metric frame, enabling later stages to attach trajectories, motions, avatars, and agent queries without additional alignment.

For avatars, we build the avatar pool $\mathcal{H}$ from two sources. First, we select 195 everyday-clothed subjects from DNA-Rendering~\cite{cheng2023dnarendering} and train HumanGS-style animatable Gaussian avatars \cite{moreau2024humangs}. These capture-based avatars provide high-quality geometry and appearance, but require roughly six hours of optimization per identity. To improve scalability, we further use GPT-Image2 to generate 305 diverse full-body human snapshots and reconstruct Gaussian avatars from them with LHM~\cite{qiu2025lhm}, taking about two minutes per identity. This image-conditioned path expands appearance diversity while remaining compatible with the same Gaussian renderer. All avatars are stored with source, scale, alignment, and generation metadata, aligned to the scene ground convention, and cached for social episode synthesis. Scene and avatar visualizations are provided in \cref{sec:assets_visualize}.

\subsection{Social episode synthesis}
\label{sec:socialgs_episode_synthesis}

Given the prepared scenes and avatar pool $\mathcal{H}$, the second stage instantiates each static environment as a social episode containing moving avatars.  We decompose this synthesis process into three steps:

\begin{enumerate}[leftmargin=*]
    \item \textbf{Semantic intent generation.}  We use Gemini 3.1 Flash to generate high-level pedestrian intents from the scene semantics, with the detailed prompt provided in Appendix~\ref{prompts}. Given the top-down semantic map, region/object labels, and traversability hints, Gemini outputs structured intents $\mathcal{I}_{\mathrm{raw}}=\{(a_i,s_i,g_i)\}_{i=1}^{M}$, where $a_i$ is a plausible activity and $(s_i,g_i)$ are its semantic start and goal locations. These intents define what pedestrians intend to do, but not their exact paths. We filter them with traversability and occupancy constraints to obtain the executable intent set $\mathcal{I}$.
    
    \item \textbf{Multi-avatar trajectory planning.}  Each intent in $\mathcal{I}$ is converted into a pedestrian root trajectory by a Recast/DetourCrowd-based planner \cite{mononen2026recastnavigation}, which inherently supports multi-agent collision avoidance among pedestrians. For avatar $i$, the planner outputs $\tau_i=\{(\mathbf{p}_{i,t},\theta_{i,t})\}_{t=1}^{T_i}$, where $\mathbf{p}_{i,t}$ and $\theta_{i,t}$ are the root position and heading. Trajectories are planned jointly, so crowded or infeasible configurations can be replanned or removed before motion synthesis.
    
    \item \textbf{Motion and avatar binding.}  The planned root trajectories determine where each pedestrian moves, but not how the body executes the movement.  Therefore, we use Kimodo~\cite{rempe2026kimodo} to synthesize full-body motion $\mathbf{Q}_i=\{\mathbf{q}_{i,t}\}_{t=1}^{T_i}$ from each trajectory $\tau_i$, conditioned on the route, heading sequence, duration, and activity descriptor $a_i$.  We then sample identities from $\mathcal{H}$ and bind their Gaussian avatars to the generated motions, producing animated avatar-Gaussian caches for each episode.
\end{enumerate}

\begin{figure}[ht!]
    \centering
    \includegraphics[width=\linewidth]{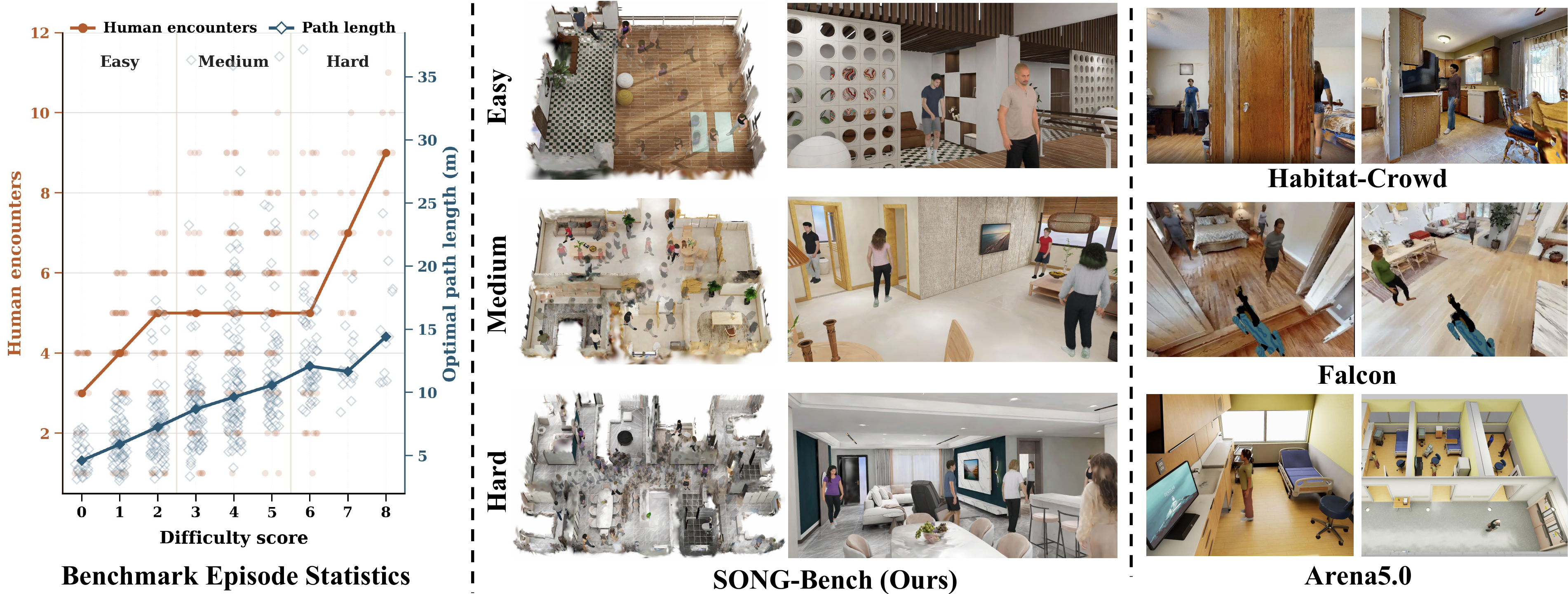}
    \caption{\textbf{SONG-Bench Overview.} Left: episode statistics across difficulty levels, measured by human avatar encounters and optimal path length. Middle: scene snapshots from SONG-Bench under easy, medium, and hard settings. Right: visual examples from other representative platforms, including Habitat-Crowd~\cite{vuong2024habicrowd}, Falcon~\cite{gong2025cognition}, and Arena5.0~\cite{kastnerarena}.}
    \label{fig:3}
\end{figure}

The final output is a synchronized episode $\mathcal{E}=\{(\tau_i,\mathbf{Q}_i,h_i)\}_{i=1}^{N}$, where each human is assigned a root trajectory, full-body motion, and Gaussian avatar.
At any timestamp, a \platform runtime can query the posed avatar Gaussians in the shared scene frame and render them together with the static scene Gaussians. Fig. \ref{fig:3} further compares \benchname with representative social platforms, highlighting its joint support for photorealistic 3DGS rendering and dynamic Gaussian avatars with natural full-body motions. Further details are provided in \cref{supp:sec:implementation_details}.

\subsection{Closed-Loop agent runtime}
\label{sec:socialgs_agent_runtime}

While photorealistic dynamic social scenes are the core of \platform, a closed-loop robot platform must also provide robot assets, physics simulation, and fine-grained collision handling.  We therefore integrate \platform with Isaac Lab~\cite{mittal2025isaaclab}, coupling its robot embodiment and physics backend with our real-time Gaussian observation layer.  At each simulator step, Isaac Lab sends the robot state, camera pose, or action query to the \platform runtime. The runtime evaluates the social episode at the queried time, composes the static scene Gaussians with the posed avatar Gaussians, and returns synchronized RGB-D observations and state signals from the robot viewpoint. Since rendering is performed on demand from the queried pose, the same episode can be reused for policy training, closed-loop rollout, simulator integration, and replay-based evaluation.

Within this runtime, collision handling is implemented through two separate paths. Static scene collisions are resolved by Isaac Lab using imported scene geometry or collision proxies. Dynamic avatar collisions are evaluated directly on the Gaussian avatars: given the robot collision volume $\mathcal{B}_t$ at time $t$, the runtime counts avatar Gaussian centers lying inside $\mathcal{B}_t$ and flags a collision when this count exceeds a robot-specific threshold $\eta_{\mathrm{col}}$. The raw count is also returned, making the criterion reproducible and adjustable across robot embodiments. This design cleanly separates scene contact from avatar contact while exposing both through a unified closed-loop query interface.

\section{Evaluation protocol}
\label{eval_protocol}
\subsection{Task definition}

\benchname focuses on the vision-based point-goal social navigation task. Given RGB-D observations and a target coordinate, an agent must navigate to the destination while avoiding collisions with static obstacles and dynamic pedestrians, and maintaining a comfortable distance from pedestrians to respect social norms. Pedestrians follow prescribed trajectories independent of the robot, reflecting the non-interactive setting in which the robot must adapt to avatar motion unilaterally.

Formally, at each timestep $t$, the agent receives a visual observation $o_t \in \mathcal{O}$ and the target position $\mathbf{g} \in \mathbb{R}^3$, and outputs a navigation action $a_t \sim \pi(o_t, \mathbf{g})$ that transitions it to a new state. Throughout the episode, we track the number of collisions with static obstacles $C_\text{obs}$, the number of collisions with avatars $C_\text{hum}$, the distance to the nearest avatar $d_\text{hum}^t$, and the distance to the goal $d_\text{goal}^t$. An episode is considered successful only if the following conditions are jointly satisfied before the step budget $T_\text{max}$ is exhausted:
\begin{equation}
d_\text{goal}^t < \delta_\text{goal}, \quad C_\text{obs} < \delta_\text{obs}, \quad C_\text{hum} = 0
\end{equation}
where $\delta_\text{goal}$ and $\delta_\text{obs}$ are the goal-reaching distance threshold and maximum tolerated static-obstacle collision count, respectively. The agent operates at a maximum linear velocity of $0.5$ m/s and $45$ °/s angularly. Pedestrian speeds are sampled between $0.5$–$1.0$ m/s. 

\subsection{Episode generation and difficulty grading}
A key requirement for a meaningful social navigation benchmark is that the robot inevitably encounters pedestrians along its route. To this end, start and goal positions for each episode are sampled within the bounding circle enclosing all pedestrian trajectories. We define four sampling modes: (1) \textbf{Co-directional}: the start is sampled within 2\,m of a pedestrian's trajectory start, and the goal within 2\,m of that pedestrian's trajectory end, aligning the robot's route with the pedestrian's direction of travel; (2) \textbf{Counter-directional}: the reverse of the above, setting the robot on a path opposing the pedestrian's direction of travel; (3) \textbf{Encounter}: start and goal are sampled freely within the bounding circle with a subtended angle greater than $90^\circ$ at the circle center to avoid trivially short paths, subject to the constraint that the robot, traversing the A$^*$ path at 0.5\,m/s, passes within 1\,m of at least one pedestrian; (4) \textbf{Random}: start and goal are sampled within the bounding circle with a subtended angle greater than $90^\circ$, without any explicit encounter constraint.

Following the synthesis method described in \cref{sec:socialgs_episode_synthesis}, we first generate 500 diverse pedestrian scenarios across 500 scenes. Testing episodes are generated via uniform sampling from the four navigation modes, resulting in a suite of 500 social navigation episodes.
We categorize the 500 evaluation episodes into easy, medium, and hard levels using four complexity factors derived from the smoothed A$^*$ path: (1) Path length (m); (2) Significant turns, i.e., the count of heading changes exceeding $40^\circ$; (3) Human avatar encounters, the number of pedestrians within 3m of the agent assuming optimal traversal at 0.5m/s; and (4) Corridor width, the mean distance to obstacles. Each factor is independently partitioned into three equal-frequency bins and assigned a score of 0, 1, or 2 (representing easy, medium, and hard). These are summed into a composite difficulty score ranging from 0 to 8. To ensure a challenging distribution, we designate the top 100 episodes as \textit{hard}, the next 200 as \textit{medium}, and the remaining 200 as \textit{easy}. In cases of tied composite scores, episodes are prioritized based on avatar encounters followed by path length. We also present the resulting benchmark statistics and select representative scenes from the three difficulty levels in Fig. \ref{fig:3}.

\subsection{Evaluation metrics}
We evaluate navigation performance across 3 dimensions: effectiveness, safety, and social compliance. 

\textbf{Effectiveness:} \textit{SR $\uparrow$} (Success Rate) measures the ratio of episodes reaching the goal. Follow \cite{anderson2018evaluation, igibsonchallenge2021}, we use \textit{SPL $\uparrow$} (Success weighted by Path Length) and \textit{STL $\uparrow$} (Success weighted by Time) to measures path efficiency, defined as $SR \cdot \frac{L^*}{\max(L, L^*)}$ and $SR \cdot \frac{T^*}{\max(T, T^*)}$ respectively, where $L^*$ and $T^*$ are the A$^*$ optimal length and time. \textit{AvgT $\downarrow$} denotes the mean completion time calculated exclusively over successful episodes. \textit{Time-out (TO $\downarrow$)} reflects the percentage of episodes exceeding the step budget.
\textbf{Safety:} \textit{CR$_{\text{obs}}$ $\downarrow$} and \textit{CR$_{\text{hum}}$ $\downarrow$} quantify failures resulting from contacts with static obstacles and dynamic pedestrians, respectively. 
\textbf{Social Compliance:} \textit{PSC $\uparrow$} (Personal Space Compliance) represents the ratio of steps maintaining a nearest-avatar distance $d_\text{hum}^t \geq 0.5$m. \textit{MinD $\uparrow$} (Minimum Human avatar Distance) tracks the closest robot-human proximity during traversal. \textbf{JK} $\downarrow$ quantifies trajectory smoothness as the RMS of the temporal curvature rate $\Delta\kappa / \Delta t$, where $\kappa = |\dot{x}\ddot{y} - \dot{y}\ddot{x}| / |\mathbf{v}|^3$, $|\mathbf{v}| = 0.5$ m/s is the agent's linear velocity, and $(x, y)$ denote the trajectory waypoint coordinates.

\section{Experiments}
\label{sec:experiments}
Our evaluation comprises two primary components:

\textbf{Zero-shot Benchmarking:} We conduct a comprehensive zero-shot evaluation of leading pre-trained baselines on our benchmark to identify performance gaps when navigating high-fidelity 3DGS-based social scenarios and diverse geometries.

\textbf{Sim-to-Real Transfer:} We fine-tune the NavDP \cite{cai2025navdp} foundation model in our platform and deploy the resulting policy onto a physical quadruped robot to verify the platform's cross-domain fidelity and the consistency of navigation performance in real-world environments.
\subsection{Zero-shot benchmarking}
\label{sec:exp_setup}
\paragraph{Setup.}In our experiments, we set $\delta_\text{goal} = 1.0$ m, $\delta_\text{obs} = 5$, and $T_\text{max} = 200$ steps. Upon a static obstacle collision, we apply a hard-coded recovery strategy: the agent is first restored to its last safe pose and $C_\text{obs}$ is incremented by one. If the agent is not stuck, determined by checking whether it has moved within the past 10 steps, execution resumes normally. If the agent is stuck, it is additionally reoriented toward the most open direction in its field of view before execution continues.

\paragraph{Baselines.} 
We select four representative state-of-the-art visual navigation methods: 
(1) \textbf{NavDP} \cite{cai2025navdp} is a navigation model trained on multi-modal goals, utilizing a diffusion-based framework to generate trajectories conditioned on ego-centric RGB-D observations and specified goals.
(2) \textbf{iPlanner} \cite{yang2023iplanner} is an iterative generative planning framework that models path generation as an imperative sequence of waypoints from depth-only measurements and goal positions, optimized for dynamic obstacle avoidance.
(3) \textbf{SocialNav} \cite{chen2511socialnav} is a socially-aware foundation model that integrates RGB images, goal coordinates, and textual prompts with human motion forecasting to ensure safety and comfort in human-populated spaces. 
(4) \textbf{NoMaD} \cite{sridhar2024nomad} is a pioneer diffusion policy trained on large-scale real-world datasets using RGB observations. Notably, as NoMaD is an image-goal navigation model, we evaluate it by rendering the RGB observation at each episode's goal position as the target input.
All baselines are tested in a zero-shot manner using their publicly released weights to examine their direct generalization capabilities.

\begin{figure}[ht!]
    \centering
    \begin{minipage}{0.48\linewidth}
        \centering
        \captionof{table}{Evaluation in avatar-free environments.}
        \label{tab:ablation_human_free}
        \resizebox{\linewidth}{!}{%
        \begin{tabular}{lcccccc}
            \toprule
            \multirow{2}{*}{Method} 
            & \multicolumn{5}{c}{Effectiveness} & \multicolumn{1}{c}{Safety} \\
            \cmidrule(lr){2-6} \cmidrule(lr){7-7}
            & SR~$\uparrow$ & SPL~$\uparrow$ & STL~$\uparrow$ & TO~$\downarrow$ & AvgT~$\downarrow$
            & CR$_{\text{obs}}$~$\downarrow$ \\
            \midrule
            NoMaD     & $39.50$ & $\mathbf{38.96}$ & $\mathbf{20.58}$ & $1.50$ & $\mathbf{18.71}$ & $\mathbf{59.00}$ \\
            NavDP     & $37.50$ & $35.42$  & $9.35$ & $\mathbf{0.00}$ & $53.40$ & $62.50$ \\
            iPlanner  & $23.00$  & $21.08$  & $12.30$ & $\mathbf{0.00}$ & $23.99$ & $77.50$ \\
            SocialNav & $\mathbf{41.00}$  & $30.68$  & $16.29$ & $\mathbf{0.00}$ & $37.43$ & $\mathbf{59.00}$ \\
            \bottomrule
        \end{tabular}%
        }
    \end{minipage}
    \hfill
    \begin{minipage}{0.48\linewidth}
        \centering
        \includegraphics[width=\linewidth]{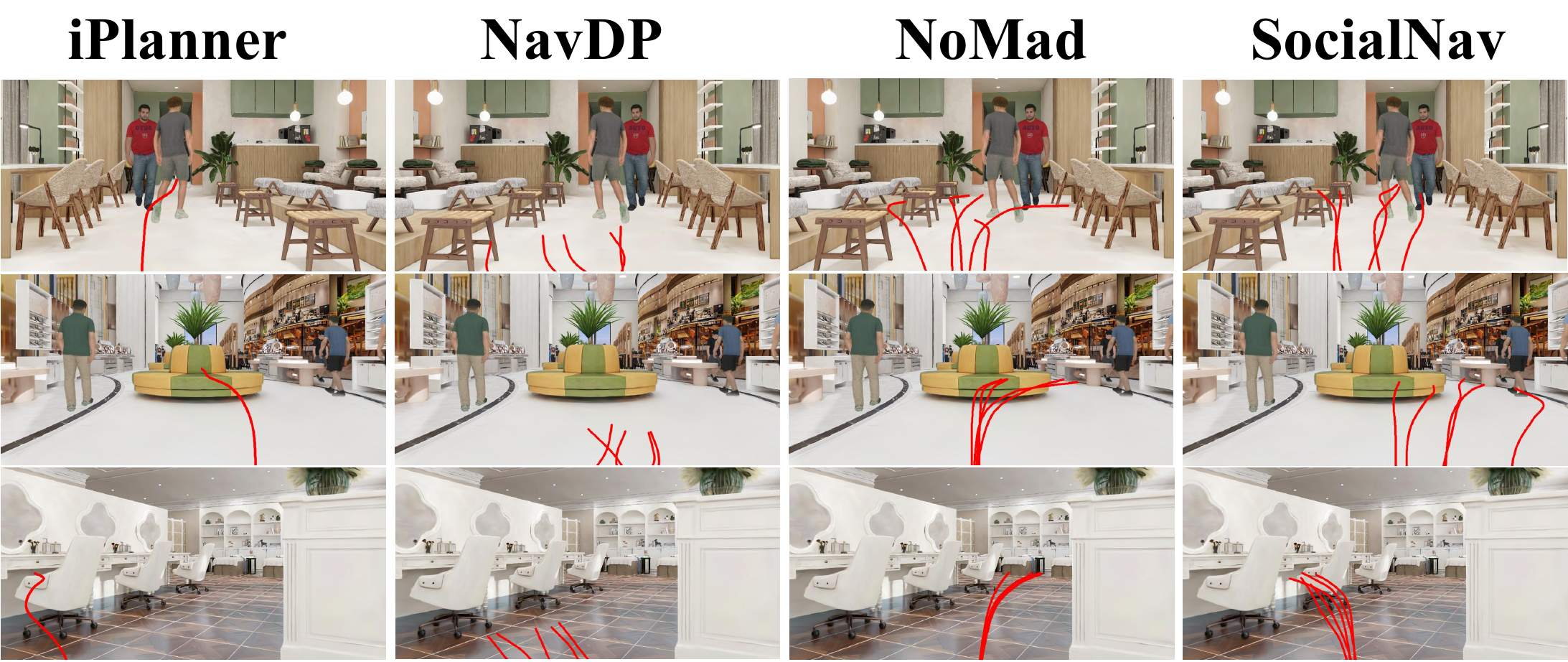}
        \caption{Visualization of navigation behaviors.}
        \label{fig:visualization}
    \end{minipage}
\end{figure}

\paragraph{Result Analysis.}
To establish a performance ceiling, we include a human-teleoperated upper bound in which 5 operators navigate using WASD controls. These operators rely solely on first-person RGB observations and relative goal positions, with no access to global maps, privileged information, or multiple attempts.
The experimental results are summarized in \cref{tab:main_results}, and several critical insights emerge from these results:

\begin{itemize}[leftmargin=*]
    \item \textbf{Vision-based Social Navigation is Far from Solved}. All baselines exhibit a poor performance drop on \benchname. The SR for all remains below 22\% in the \textit{Easy} set and plummets to near-zero in the \textit{Hard} set.  The experimental results reveal a stark reality: despite significant progress in previous simulators, current state-of-the-art methods are still far from achieving reliable social navigation in high-fidelity environments. This collective performance gap underscores that photorealistic visual cues and complex human dynamics introduce challenges that existing architectures are yet to overcome, positioning \platform as a critical milestone for future research.
    \item \textbf{Safety Deficit Preceding Social Etiquette}. Our results reveal that current models require significant improvements in safety performance before considering complex social etiquette. Specifically, all four baselines exhibit failure rates exceeding $75\%$ in the \textit{Easy} scenarios due to either human avatar or obstacle collisions (NoMaD $\sim$ $76.5\%$, NavDP $\sim$ $87
    .0\%$, iPlanner $\sim$ $85.0\%$, and SocialNav $\sim$ $81.0\%$). This underscores that vision-based policies still struggle with reliable obstacle avoidance in unfamiliar scenes. Our ablation study in avatar-free environments (\cref{tab:ablation_human_free}) confirms this, as high failure rates persist even without dynamic avatars, revealing the inherent difficulty of mapping photorealistic observations to safe geometric spaces. Furthermore, current models exhibit a severe lack of collision recovery mechanisms. Once a collision occurs, agents typically remain stuck rather than re-planning or backing out. This bottleneck stems from the prevailing reliance on perfect expert demonstrations, which fail to provide the "error-and-recovery" transitions essential for handling real-world contact scenarios.
    \item \textbf{Real-World Data Distribution Matters More Than Model Scale.} Among NoMaD, NavDP, and SocialNav, all of which adopt expert trajectory-based imitation learning, NoMaD consistently outperforms the others across all difficulty levels despite its comparatively simpler architecture. We attribute this to training data distribution: NoMaD is trained on over 100 hours of real-world trajectories, NavDP relies entirely on simulation, and SocialNav combines both. This ranking suggests that \platform's photorealistic 3DGS rendering and LLM-driven pedestrian behaviors produce observations that more closely mirror real-world distributions than conventional simulators, both in visual fidelity and human dynamics. Notably, when evaluated in human-free environments (Table~\ref{tab:ablation_human_free}), the performance gap among these three methods largely vanishes. This reveals that the ranking observed in the main benchmark stems not from differences in scene geometry understanding, but from each model's capacity to handle dynamic human agents. NoMaD's real-world training implicitly captures authentic pedestrian patterns that simulation fails to replicate, underscoring dynamic human interaction as the critical bottleneck and positioning \platform as a uniquely discriminative testbed for evaluating this capability.
\end{itemize}

\paragraph{Qualitative results}
\cref{fig:visualization} shows the predicted trajectory of different baselines across three scenarios. iPlanner follows a deterministic policy, while the other three baselines utilize generative policies.

\begin{table*}[t]
    \centering
    \caption{\textbf{Zero-shot benchmark results on \benchname across three difficulty levels.} The best results among all learning-based baselines are highlighted with \textbf{bold} colored backgrounds.}
    \label{tab:main_results}
    \resizebox{\linewidth}{!}{%
    \begin{tabular}{llcccccccccc}
        \toprule
        \multirow{2}{*}{Difficulty} & \multirow{2}{*}{Method} 
        & \multicolumn{5}{c}{Effectiveness} & \multicolumn{2}{c}{Safety} & \multicolumn{3}{c}{Social Compliance} \\
        \cmidrule(lr){3-7} \cmidrule(lr){8-9} \cmidrule(lr){10-12}
        & & SR~$\uparrow$ (\%) & SPL~$\uparrow$ (\%) & STL~$\uparrow$ (\%) & TO~$\downarrow$ (\%) & AvgT~$\downarrow$ (s)
        & CR$_{\text{obs}}$~$\downarrow$ (\%) & CR$_{\text{hum}}$~$\downarrow$ (\%)
        & PSC~$\uparrow$ (\%) & MinD~$\uparrow$ (m) & JK~$\downarrow$ (m$^{-1}$s$^{-1}$)\\
        \midrule
        \multirow{5}{*}{Easy} 
        & NoMaD \cite{sridhar2024nomad}        
        & \easybest{21.50} & \easybest{20.94} & \easybest{9.16} & $2.00$ & \easybest{29.99} 
        & $73.50$ & \easybest{3.00} & $92.87$ & \easybest{1.05} & \easybest{0.24} \\
        & NavDP \cite{cai2025navdp}        
        & $12.50$ & $10.96$ & $2.63$ & $0.50$ & $79.45$ 
        & $39.50$ & $47.50$ & \easybest{96.86} & $0.71$ & $0.72$ \\
        & iPlanner \cite{yang2023iplanner}     
        & $15.00$ & $13.09$ & $3.18$ & \easybest{0.00} & $58.36$ 
        & $40.00$ & $45.00$ & $87.61$ & $0.76$ & $3.83$ \\
        & SocialNav \cite{chen2511socialnav}    
        & $19.00$ & $13.71$ & $5.61$ & \easybest{0.00} & $50.96$ 
        & \easybest{27.50} & $53.50$ & $87.52$ & $0.71$ & $19.61$ \\
        \cmidrule(l){2-12}
        & Human teleop 
        & $95.5$ & $85.53$ & $83.80$ & $0.00$ & $34.76$ 
        & $2.00$ & $2.50$ & $90.82$ & $0.80$ & $1.65$ \\
        \midrule\midrule
        \multirow{5}{*}{Medium} 
        & NoMaD \cite{sridhar2024nomad}       
        & \mediumbest{11.00} & \mediumbest{10.96} & \mediumbest{3.69} & $1.50$ & \mediumbest{92.41} 
        & $84.50$ & \mediumbest{3.00} & $90.69$ & \mediumbest{1.11} & \mediumbest{0.13} \\
        & NavDP \cite{cai2025navdp}       
        & $7.00$ & $6.44$ & $1.28$ & \mediumbest{0.00} & $120.88$ 
        & $50.00$ & $43.00$ & \mediumbest{93.96} & $0.45$ & $0.66$ \\
        & iPlanner \cite{yang2023iplanner}    
        & $4.50$ & $4.21$ & $0.96$ & \mediumbest{0.00} & $100.01$ 
        & $53.00$ & $42.50$ & $89.13$ & $0.78$ & $1.99$ \\
        & SocialNav \cite{chen2511socialnav}   
        & $6.00$ & $4.47$ & $1.62$ & \mediumbest{0.00} & $99.31$ 
        & \mediumbest{49.50} & $44.50$ & $88.89$ & $0.72$ & $21.46$ \\
        \cmidrule(l){2-12}
        & Human teleop 
        & $85.50$ & $74.54$ & $79.14$ & $0.00$ & $44.33$ 
        & $2.00$ & $12.50$ & $90.49$ & $0.61$ & $4.34$ \\
        \midrule\midrule
        \multirow{5}{*}{Hard} 
        & NoMaD \cite{sridhar2024nomad}       
        & \hardbest{3.00} & \hardbest{3.00} & \hardbest{1.28} & $1.00$ & $62.78$ 
        & $96.00$ & \hardbest{0.00} & $94.38$ & \hardbest{1.13} & \hardbest{0.10} \\
        & NavDP \cite{cai2025navdp}       
        & \hardbest{3.00} & $2.92$ & $0.60$ & \hardbest{0.00} & \hardbest{56.29} 
        & $63.00$ & $34.00$ & \hardbest{96.33} & $0.46$ & $0.51$ \\
        & iPlanner \cite{yang2023iplanner}    
        & \hardbest{3.00} & \hardbest{3.00} & $0.78$ & \hardbest{0.00} & $103.01$ 
        & $68.00$ & $29.00$ & $86.67$ & $0.73$ & $0.56$ \\
        & SocialNav \cite{chen2511socialnav}   
        & $0.00$ & $0.00$ & $0.00$ & \hardbest{0.00} & $-$ 
        & \hardbest{52.00} & $48.00$ & $85.67$ & $0.73$ & $9.13$ \\
        \cmidrule(l){2-12}
        & Human teleop 
        & $80.00$ & $69.29$ & $64.40$ & $0.00$ & $61.50$ 
        & $1.50$ & $18.50$ & $85.19$ & $0.52$ & $7.37$ \\
        \bottomrule
    \end{tabular}%
    }
\end{table*}

\subsection{Fine-tuning and real-world transfer}
\paragraph{Fine-tuning NavDP.}
We fine-tune NavDP using a curriculum data collection strategy. The training data consists of two primary sources: human teleoperated trajectories and successful episodes generated by a fine-tuned Height model \cite{liu2025height}. Specifically, we first fine-tune the Height planner on 2D semantic maps using 5k episodes across 500 diverse scenarios. During this stage, the model is provided with rich spatial priors, including the precise positions of all pedestrians and obstacles, achieving a $75\%$ success rate on these 5k training episodes. We then select all high quality trajectories from these successful episodes along with human teleoperated data for replaying within \platform. This process allows us to collect synchronized multi-modal data for a total of approximately 2k trajectories, which are subsequently used to fine-tune the vision based NavDP policy.

\begin{figure}[ht!]
    \centering
    \includegraphics[width=\linewidth]{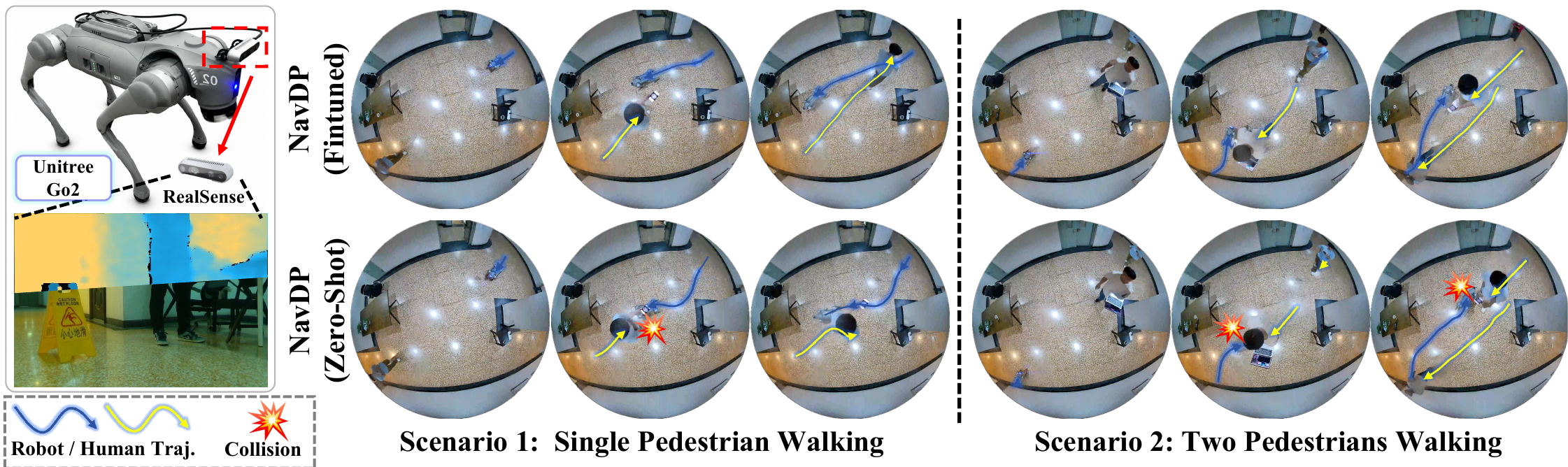}
    \caption{\textbf{Real-World Experiment Visualization.} The left part illustrates the real-world robotic platform and sensor setup, while the right part shows navigation trajectories and collision outcomes of NavDP under fine-tuned and zero-shot settings in one-person and two-person walking scenarios.}
    \label{real-world}
\end{figure}

\paragraph{Real-world experimental setup.}
As shown in \cref{real-world}, we conduct real world experiments to verify whether the high quality multi-modal data curated from \benchname can effectively improve the generalization capability of navigation policies. We utilize a Unitree Go2 robot equipped with an Intel Realsense D435i camera to compare the performance of the zero-shot and fine-tuned NavDP. For all trials, the robot is assigned a goal position 7 m directly ahead, with state estimation relying solely on the onboard odometry provided by the Go2 platform. We define two social settings: (1) a single pedestrian scenario where one person walks towards the robot; and (2) a two pedestrian scenario where two people approach in a staggered, one-after-another formation. Each setting consists of 20 trials to examine the model's ability to handle dynamic social conflicts.

\paragraph{Transfer Results.}
\begin{wraptable}[6]{r}{0.35\textwidth}
    \centering
    \small
    \vspace{-6pt}
    \captionof{table}{\textbf{SR of NavDP under real world transfer.}}
    \label{tab:transfer_results}
    \resizebox{\linewidth}{!}{%
    \begin{tabular}{lcc}%
        \toprule
        \textbf{Setting}  & \textbf{Zero-shot} & \textbf{Fine-tuned} \\ 
        \midrule
        Single Ped.       & 10\%               & \textbf{20\%}       \\
        Two Peds.         & 5\%                & \textbf{10\%}       \\ 
        \bottomrule
    \end{tabular}%
    }%
\end{wraptable}
As shown in Table \ref{tab:transfer_results}, fine-tuning on \benchname improves the success rate of NavDP in both real world settings. Specifically, the success rate increases from 10\% to 20\% in the single pedestrian scenario and from 5\% to 10\% in the two pedestrian case. While absolute performance remains limited, these improvements indicate that our high quality multi-modal data helps the model better navigate in physical environments.

\section{Conclusion}
We presented \platform, a photorealistic social navigation benchmark characterized by 3DGS scenes and avatars, semantically driven pedestrian trajectories, and natural motion synthesis. Building upon this, \benchname provides rich evaluation episodes and comprehensive metrics for vision-based social navigation. Evaluation of representative baselines reveals the task remains far from solved, with current models lacking essential obstacle avoidance and collision recovery capabilities. We also find data realism trumps model scale. We believe \platform and \benchname will significantly facilitate future research in developing robust social navigation agents.
\bibliographystyle{plain}
\bibliography{reference}

\newpage

\appendix
\crefalias{section}{appendix}
\renewcommand\thefigure{A\arabic{figure}}
\setcounter{figure}{0}
\renewcommand\thetable{A\arabic{table}}
\setcounter{table}{0}
\renewcommand\theequation{A\arabic{equation}}
\setcounter{equation}{0}

\section*{Appendix}

\section{Assets visualization}\label{sec:assets_visualize}

\begin{figure}[p]
    \centering
    \includegraphics[width=\linewidth]{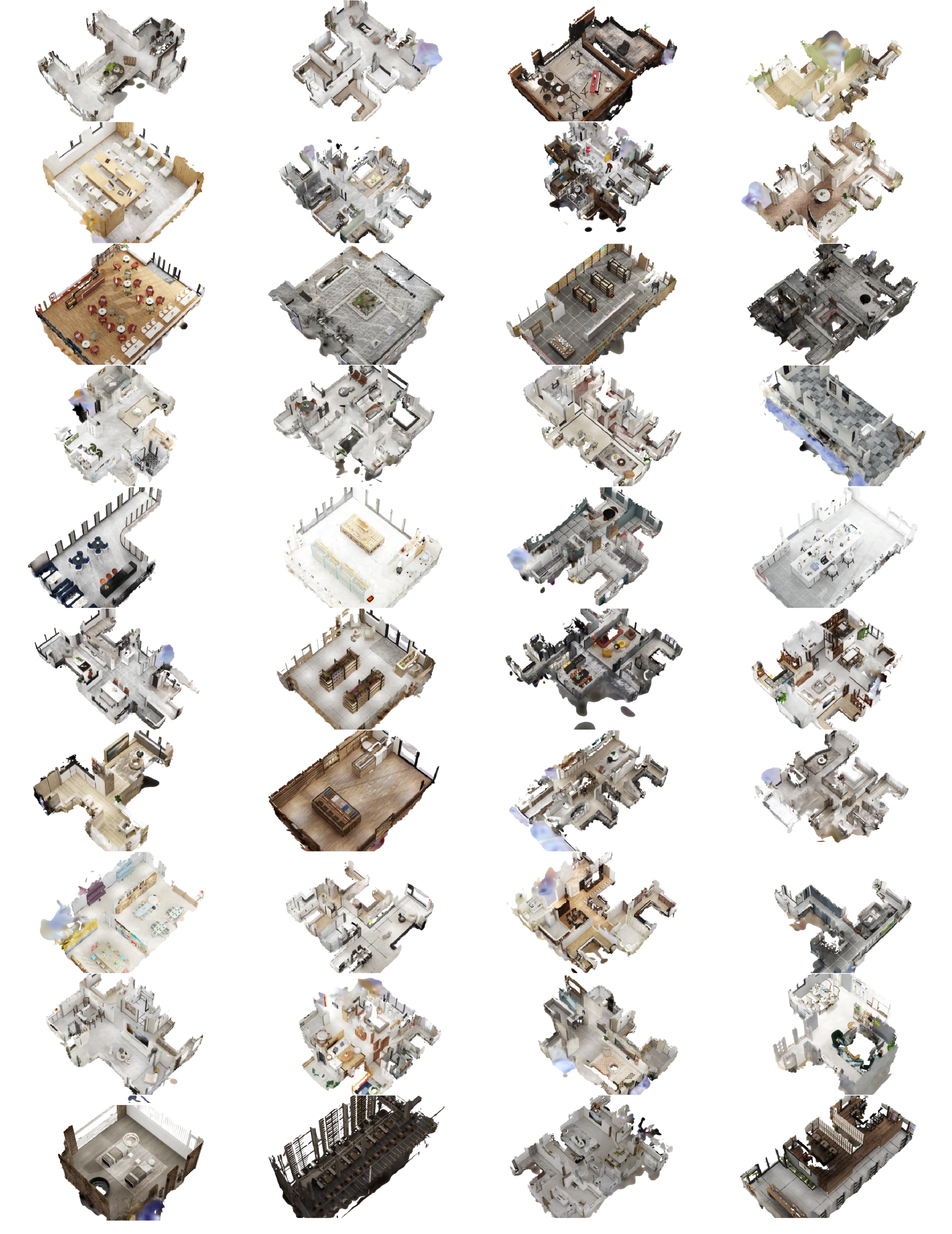}
    \caption{Snapshots of Our 3DGS Scenes.}
    \label{fig:Snapshots1}
\end{figure}

\begin{figure}[p]
    \centering
    \includegraphics[width=\linewidth]{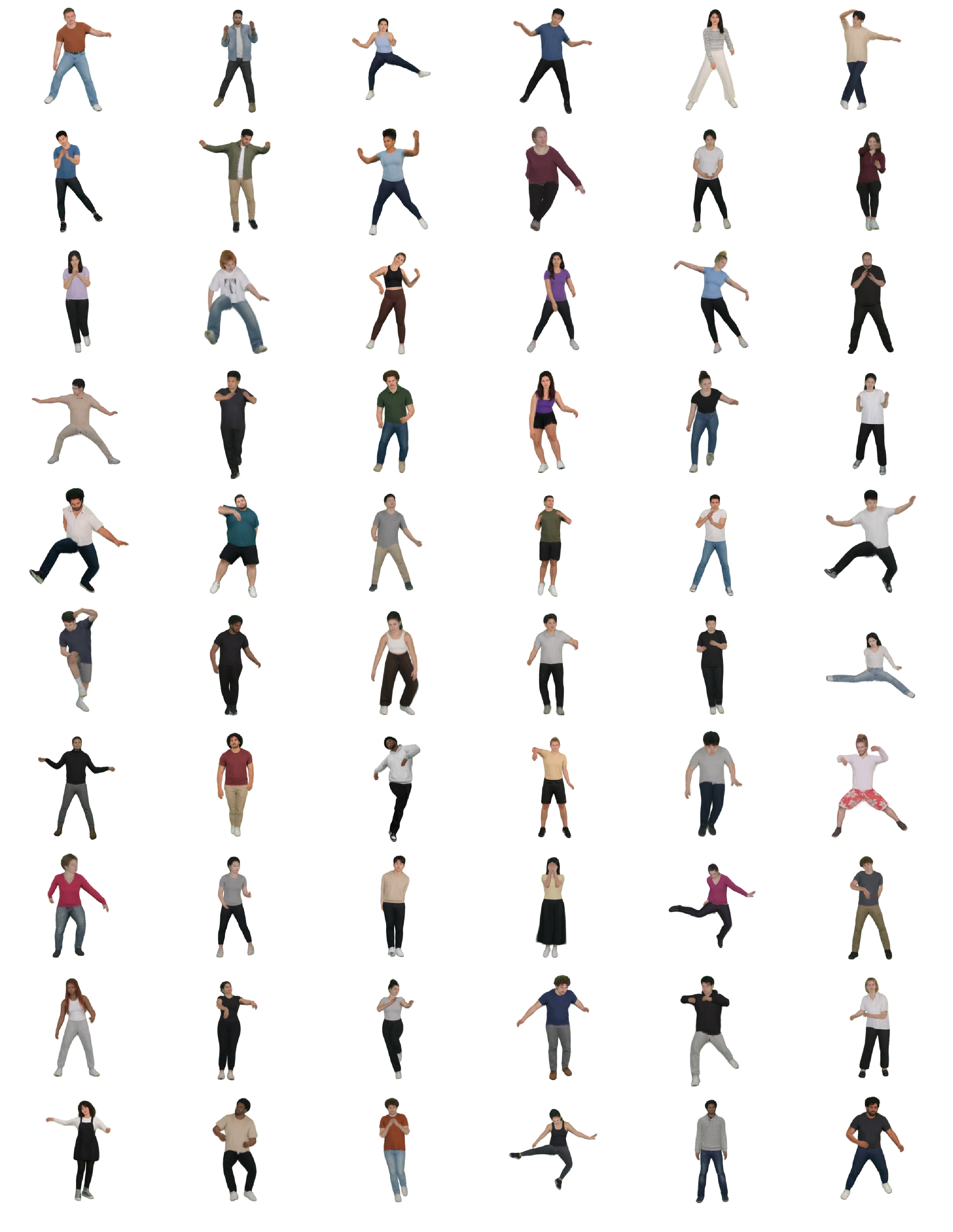}
    \caption{Snapshots of Our HumanGS Avatars with Randomly Sampled Motions.}
    \label{fig:Snapshots2}
\end{figure}

\begin{figure}[p]
    \centering
    \includegraphics[width=\linewidth]{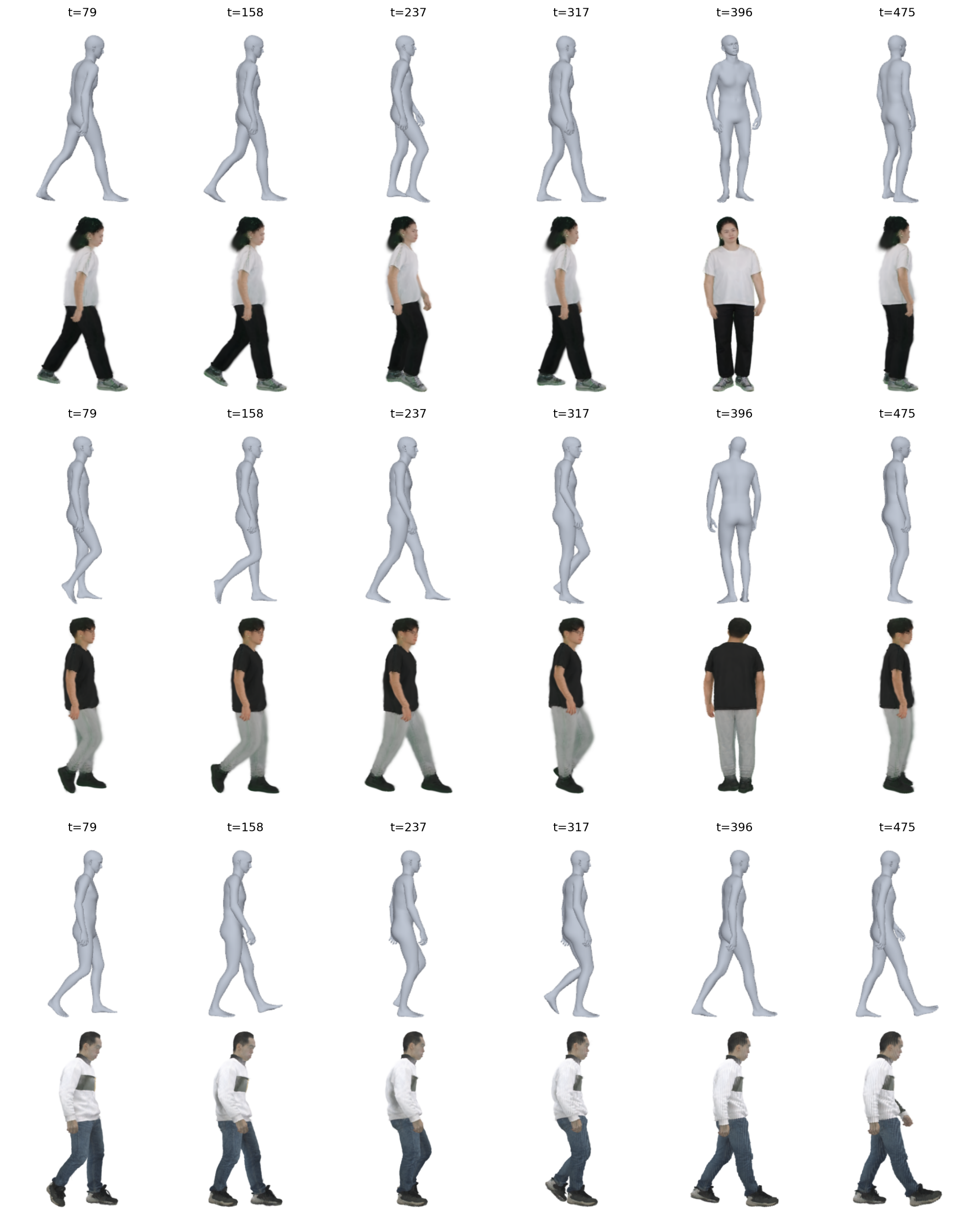}
    \caption{Temporal Snapshots of Our HumanGS Avatars along Motion Sequences.}
    \label{fig:Snapshots3}
\end{figure}

\paragraph{Asset examples.}
\platform integrates static 3DGS scene assets sourced from SAGE-3D~\cite{miao2025sage3d} with animatable HumanGS avatars to instantiate photorealistic dynamic social-navigation episodes. 
\cref{fig:Snapshots1} presents representative SAGE-3D 3DGS scenes used as static environments, while \cref{fig:Snapshots2} shows diverse HumanGS avatars under randomly sampled motions. 
\cref{fig:Snapshots3} further visualizes motion-driven temporal sequences, where the HumanGS avatars are animated by generated full-body motions from Kimodo.

\section{Implementation Details}
\label{supp:sec:implementation_details}

\paragraph{Platform construction details.}
For all experiments, each adopted SAGE-3D-style scene is organized as a unified scene package containing \texttt{3dgs\_compressed.ply}, \texttt{scene\_flat.usda}, and \texttt{semantic\_map.json}. The semantic map is used only for offline Social Episode Synthesis, while all agent observations are rendered directly from the Gaussian scene. Pedestrian root trajectories are generated at 30 FPS and stored as deterministic episode caches. During evaluation, the Closed-Loop Agent Runtime queries these caches according to the simulator timestep. For cached Gaussian avatar replay, we store per-frame Gaussian positions and rotations as FP16 arrays, together with static Gaussian attributes and root-alignment metadata. The Gaussian runtime uses a $70^\circ$ horizontal field of view, near/far planes of $0.01\,\mathrm{m}$ and $100\,\mathrm{m}$, and a default rendering resolution of $960{\times}540$. The robot camera is mounted $0.8\,\mathrm{m}$ above the ground. For dynamic avatar collision checking, the robot is approximated by a collision volume centered at the camera/body height with a radius of $0.2\,\mathrm{m}$, and an avatar collision is reported when the number of avatar Gaussian centers inside the volume exceeds the configured threshold $\eta_{\mathrm{col}}=5$.

\paragraph{Kimodo motion generation details.}
For each planned pedestrian trajectory, we use Kimodo to synthesize trajectory-conditioned full-body motion. The text prompt provided to Kimodo is \textit{``A person is casually walking forward at a natural pace with minimal waist sway''}, which encourages stable and natural walking motions without exaggerated body oscillations. Although Kimodo generates plausible full-body motion from the planned route, the generated root motion is not always exactly aligned with the Recast/DetourCrowd trajectory used by the social episode. Therefore, before avatar binding, we apply a trajectory-consistency post-processing stage to each Kimodo output. Specifically, we first align the generated AMASS-format root translation and heading with the planned trajectory in the shared scene coordinate frame. The motion sequence is then temporally resampled to match the trajectory timestamps and smoothed to reduce frame-level jitter. We further apply a foot-ground correction pass to mitigate floating and foot-sliding artifacts, ensuring that the final motion remains physically plausible when replayed by the Gaussian avatar. For reproducibility, both the post-processed Kimodo motion and the converted AMASS file are retained. The processed motion files are stored under \texttt{human\_amass/traj\_\{id\}/} and serve as the only motion inputs to the HumanGS cache builder for subsequent avatar-Gaussian replay.

\paragraph{Evaluation and fine-tuning details.}
All baselines are evaluated with the same Closed-Loop Agent Runtime. Each episode runs for at most 200 steps at 10 Hz, with a goal-reaching threshold of $1.0\,\mathrm{m}$. The robot moves with a maximum linear speed of $0.5\,\mathrm{m/s}$ and angular speed of $45^\circ/\mathrm{s}$, while pedestrian speeds are sampled from $[0.5,1.0]\,\mathrm{m/s}$. Static-scene failure is triggered after 5 accumulated obstacle collisions, and human-avatar failure is triggered by any avatar collision. SPL and STL use the precomputed A$^*$ path length and the corresponding nominal traversal time under $0.5\,\mathrm{m/s}$ traversal. Personal-space compliance is computed with a $0.5\,\mathrm{m}$ nearest-avatar distance threshold, and other trajectory-level metrics are computed from the logged robot trajectory. For sim-to-real transfer, NavDP is fine-tuned without changing its architecture, observation interface, or training objective, using approximately 2K replayed trajectories collected in \platform from successful Height-planner rollouts and human teleoperation.


\section{Limitations}
\label{Limitations}

Although \platform provides a unified platform that combines photorealistic 3DGS assets, semantics-driven social episode synthesis, and closed-loop robot evaluation, it still adopts several practical simplifications. First, dynamic avatar collision is detected by counting the Gaussian centers that fall inside the robot's collision volume. This design is lightweight, reproducible, and directly compatible with animatable Gaussian avatars, but it remains an approximation of true body-level physical contact. Since Gaussian density may vary across avatars, clothing regions, and reconstruction sources, a fixed threshold may not perfectly reflect the exact contact boundary in all cases. In the current benchmark, this approximation is mainly used as a consistent safety signal for comparing navigation policies, rather than as a full physical interaction model. Future versions can improve this component by constructing avatar-specific collision proxies, fitting articulated body volumes from posed avatars, or combining Gaussian-level occupancy with mesh- or skeleton-based contact estimation. These extensions would preserve the rendering fidelity of Gaussian avatars while providing more physically grounded contact supervision.

Second, pedestrians in the current \benchname follow prescribed trajectories and do not react to the robot online. This non-interactive setting makes the benchmark deterministic and ensures that different navigation policies are evaluated under identical social scenarios, which is important for fair comparison. However, it does not fully capture reciprocal human--robot interaction, where pedestrians may slow down, yield, detour, or change their intent in response to the robot. Therefore, \benchname currently focuses on unilateral robot adaptation to dynamic pedestrians, especially the ability to perceive human motion, avoid collisions, and maintain personal space from onboard visual observations. Our experiments already show that existing methods struggle even under this controlled setting, with high failure rates persisting in both social and avatar-free environments. This suggests that robust geometric safety remains a prerequisite before more complex interactive social behaviors can be reliably evaluated. A natural next step is to incorporate reactive pedestrian policies, such as learned human response models or social-force-based local adaptation, while retaining the current fixed-trajectory protocol as a standardized diagnostic benchmark.

\section{Prompts}
\label{prompts}

\begin{figure}[ht!]
    \centering
    \includegraphics[width=\linewidth]{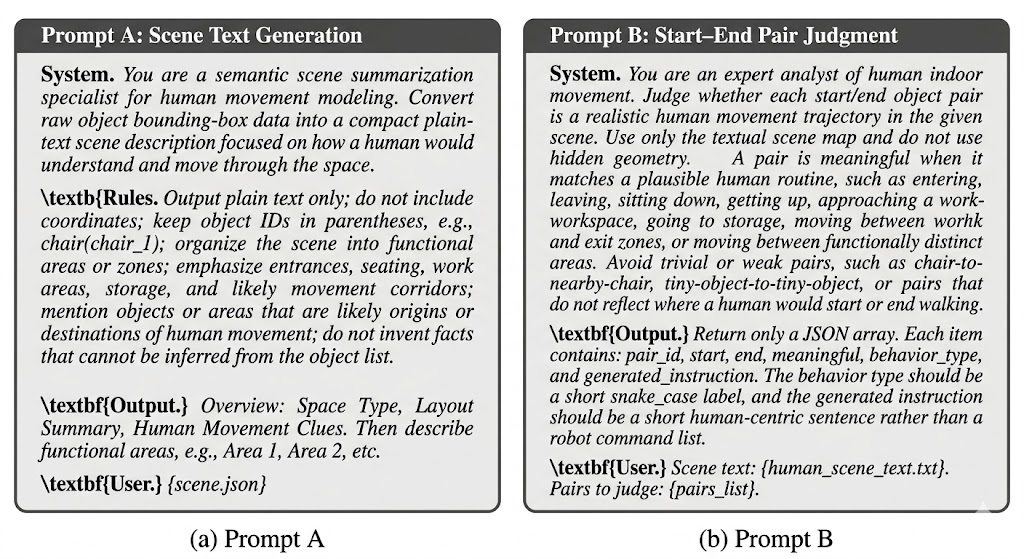}
    \caption{\textbf{Prompts for semantic social episode synthesis.} Prompt A generates a compact human-centric textual description from scene semantics, while Prompt B evaluates meaningful start--end movement pairs for pedestrian intent generation.}
    \label{fig:prompt}
\end{figure}

To synthesize semantically plausible pedestrian behaviors, we employ a two-stage prompting pipeline with Gemini 3.1 Flash, as illustrated in Fig.~\ref{fig:prompt}. 
Prompt A first converts the scene semantic map and object annotations into a compact human-centric scene description, summarizing functional areas, traversable regions, and likely movement patterns. Based on this textual scene representation, Prompt B further evaluates candidate start--end object pairs and selects meaningful human movement routines, such as entering workspaces, moving between functional regions, or accessing storage and seating areas. The resulting semantic intents are then used for the trajectory planning stage described in \cref{sec:socialgs_episode_synthesis}.

\section{Platform Scalability and Runtime Efficiency}
\label{app:platform_efficiency}

\paragraph{Profiling protocol.}
We profile \platform from two aspects: Social Episode Synthesis and Closed-Loop Agent Runtime efficiency.
The corresponding statistics are summarized in \cref{tab:episode_synthesis_efficiency,tab:runtime_efficiency}.
For Social Episode Synthesis, timing is aggregated from the selected top-500 generated episodes.
For runtime profiling, we sample 50 scenes from \benchname, with approximately three active avatars per scene on average.
All runtime results in \cref{tab:runtime_efficiency} are measured on a single NVIDIA RTX 4090 GPU.
\cref{tab:episode_synthesis_efficiency} reports the one-time cost of instantiating a static scene as a dynamic social episode, including semantic intent generation, intent filtering, multi-avatar trajectory planning, motion synthesis with Kimodo, and motion and avatar binding.
\cref{tab:runtime_efficiency} reports the online speed of the Gaussian runtime and the Closed-Loop Agent Runtime.

\paragraph{Runtime settings.}
The three runtime settings in \cref{tab:runtime_efficiency} separate standalone rendering, in-loop rendering, and the full agent-environment step.
\emph{Gaussian runtime rendering} measures RGB-D rendering from the Gaussian runtime without Isaac Lab.
\emph{In-loop Gaussian rendering} measures only the Gaussian rendering/query component when the robot camera pose is provided by the Closed-Loop Agent Runtime; it excludes the full Isaac Lab robot step, policy inference, and logging overhead.
\emph{Full Closed-Loop Agent Runtime} measures the complete environment step, including Isaac Lab robot stepping, static-scene collision, Gaussian RGB-D rendering, avatar collision queries, and metric logging.
The released code provides implementation-level documentation for these stages, including executable scripts and YAML/JSON configuration files for social episode synthesis, cached replay, Gaussian rendering, Isaac Lab integration, collision queries, logging, and metric computation.

\paragraph{Cache-based replay.}
All official \benchname episodes are stored as deterministic caches.
Once an episode is generated, evaluation does not require re-running Gemini, Recast/DetourCrowd, or Kimodo.
Therefore, the Social Episode Synthesis cost in \cref{tab:episode_synthesis_efficiency} is a one-time preprocessing cost rather than a per-evaluation overhead.
During evaluation, the Closed-Loop Agent Runtime directly replays the cached dynamic avatars and renders synchronized RGB-D observations from the current robot viewpoint, as profiled in \cref{tab:runtime_efficiency}.

\begin{table*}[t!]
    \centering
    \small
    \setlength{\tabcolsep}{5pt}
    \renewcommand{\arraystretch}{1.12}
    \caption{Efficiency of Social Episode Synthesis. Timings are aggregated from the selected top-500 generated episodes. Values marked with $\approx$ are estimated because Gemini/API timing and intent filtering were not isolated in the old logs.}
    \label{tab:episode_synthesis_efficiency}
    \begin{tabular}{@{}lccc@{}}
    \toprule
    \textbf{Stage} & \textbf{Unit} & \textbf{Time} & \textbf{Device} \\
    \midrule
    Semantic intent generation & per scene & $\approx$15 s & API/CPU \\
    Intent filtering & per scene & $\approx$1 s & CPU \\
    Multi-avatar trajectory planning & per episode & 7.86 s & CPU \\
    Motion synthesis with Kimodo & per human & 27.27 s & GPU \\
    Motion and avatar binding & per episode & 133.37 s & GPU/CPU \\
    Episode cache writing & per episode & included above & CPU \\
    \midrule
    \textbf{Total Social Episode Synthesis} & per episode & $\approx$5.3 min & -- \\
    \bottomrule
    \end{tabular}
\end{table*}

\begin{table*}[t!]
    \centering
    \small
    \setlength{\tabcolsep}{6pt}
    \renewcommand{\arraystretch}{1.12}
    \caption{Efficiency of Gaussian runtime rendering and Closed-Loop Agent Runtime. Runtime profiling is conducted on a single NVIDIA RTX 4090 GPU using 50 sampled scenes with approximately three active avatars per scene on average.}
    \label{tab:runtime_efficiency}
    \begin{tabular}{@{}lccc@{}}
    \toprule
    \textbf{Setting} & \textbf{Isaac Lab} & \textbf{Resolution} & \textbf{Throughput} \\
    \midrule
    Gaussian runtime rendering & No & $960{\times}540$ & 41.77 FPS \\
    In-loop Gaussian rendering & Yes & $854{\times}480$ & 17.31 FPS \\
    Full Closed-Loop Agent Runtime & Yes & $854{\times}480$ & 10.01 steps/s \\
    \bottomrule
    \end{tabular}
\end{table*}

\clearpage

\end{document}

%% file: config.tex
\usepackage[utf8]{inputenc}
\usepackage[T1]{fontenc}
\usepackage{latexsym}
\usepackage{microtype}
\usepackage{booktabs}
\usepackage{amsfonts}
\usepackage{graphicx}
\usepackage{algorithmic}
\usepackage[linesnumbered,ruled,vlined]{algorithm2e}
\usepackage{acronym}
\usepackage{balance}
\usepackage[skip=3pt]{caption}
\usepackage[skip=3pt]{subcaption}
\usepackage{xspace}
\usepackage{setspace}
\usepackage{tabularx,colortbl,multirow,array,makecell}
\usepackage{overpic,wrapfig}
\usepackage{nicefrac}
\usepackage[misc]{ifsym}
\usepackage{siunitx}
\usepackage[dvipsnames,svgnames,x11names]{xcolor}
\usepackage{soul} 
\usepackage{pifont}  
\usepackage[pagebackref,breaklinks,colorlinks,citecolor=cvprblue]{hyperref}
\usepackage[capitalize]{cleveref}
\usepackage{lipsum}
\usepackage{bm}
\usepackage{xcolor}
\usepackage{enumitem}

\acrodef{eai}[EAI]{Embodied AI}
\acrodef{s2r}[Sim2Real]{sim-to-real}
\acrodef{vln}[VLN]{vision-language navigation}
\acrodef{mcmc}[MCMC]{Markov-Chain Monte-Carlo}
\acrodef{cd}[CD]{Chamfer Distance}
\acrodef{ecd}[ECD]{Enhanced Chamfer Distance}
\acrodef{bbox}[Bbox]{Bounding box}
\acrodef{iou}[IoU]{Intersection over Union}
\acrodef{agv}[AGV]{Automated Guided Vehicle}
\acrodef{dwa}[DWA]{Dynamic Window Approach}

\definecolor{WeiColor}{rgb}{1,0.33,0.64}
\definecolor{myDarkBlue}{RGB}{55,81,139}
\definecolor{myLightBlue}{RGB}{158,186,217}
\definecolor{myLightGreen}{RGB}{126,171,85}
\definecolor{myRed}{RGB}{176,36,24}
\definecolor{JxColor}{RGB}{50,102,204}
\definecolor{LightGray}{gray}{0.9}
\definecolor{cvprblue}{rgb}{0.21,0.49,0.74}
\definecolor{softlime}{RGB}{180, 229, 162}

\definecolor{pointcolor}{RGB}{233,113,50}   
\definecolor{objectcolor}{RGB}{15,158,213}  
\definecolor{imagecolor}{RGB}{78,167,46}    

\makeatletter
\DeclareRobustCommand\onedot{\futurelet\@let@token\@onedot}
\def\@onedot{\ifx\@let@token.\else.\null\fi\xspace}

\makeatother

\makeatletter
\renewcommand{\paragraph}{%
  \@startsection{paragraph}{4}%
  {\z@}{0ex \@plus 0ex \@minus 0ex}{-1em}%
  {\hskip\parindent\normalfont\normalsize\bfseries}%
}
\makeatother

\newcommand{\platform}{\textbf{SONG}\xspace}
\newcommand{\benchname}{\textbf{SONG-Bench}\xspace}